\documentclass{article}

\usepackage[acronym,shortcuts,smallcaps,nowarn,nohypertypes={acronym,notation}]{glossaries}

\usepackage{microtype}
\usepackage{graphicx}
\usepackage{subfigure}
\usepackage{booktabs} % for professional tables
\usepackage{amsfonts}
\usepackage{amssymb}
\usepackage{amsmath}
\usepackage{amsthm}
\usepackage{enumitem}
\usepackage{soul}
\usepackage{bbm}

\newacronym{ood}{ood}{out-of-distribution}
\newacronym{dgm}{dgm}{deep generative model}
\newacronym{mnist}{mnist}{Modified National Institute of Standards and Technology database}
\newacronym{svhn}{svhn}{svhn}
\newacronym{cnn}{cnn}{cnn}
\newacronym{pixelcnn++}{pixelcnn++}{pixelcnn++}
\newacronym{pixelcnn++v2}{pixelcnn-v2}{pixelcnn-v2}
\newacronym{glow}{glow}{glow}
\newacronym{cifar-10}{cifar-10}{cifar-10}
\newacronym{auc}{auc}{area under the curve}
\newacronym{roc}{roc}{receiver operating characteristic}
\newacronym{oos}{oos}{out-of-support}
\newacronym{mle}{mle}{maximum likelihood estimation}

\newacronym{bpd}{bpd}{bits-per-dimension}

\DeclareRobustCommand{\indicator}[1]{\ensuremath{\mathbbm{1}\left[#1\right]}}

\newcommand{\mbtheta}{\theta}
\newcommand{\mbx}{\textbf{x}}
\newcommand{\mby}{\textbf{y}}
\newcommand{\E}{\mathbb{E}}
\newcommand{\g}{\,\vert\,}
\newcommand{\minus}{\scalebox{0.75}[1.0]{$-$}}

\newtheorem{proposition}{Proposition}

\usepackage{hyperref}
\usepackage{units}
\usepackage{array}
\newcolumntype{?}{!{\vrule width 1pt}}

\usepackage[nameinlink]{cleveref}

\usepackage[accepted]{icml2021}

\icmltitlerunning{Understanding Failures in Out-of-Distribution Detection with Deep Generative Models}

\begin{document}

\twocolumn[
\icmltitle{Understanding Failures in Out-of-Distribution Detection with \\Deep Generative Models}

\begin{icmlauthorlist}
\icmlauthor{Lily H. Zhang}{nyu}
\icmlauthor{Mark Goldstein}{nyu}
\icmlauthor{Rajesh Ranganath}{nyu}
\end{icmlauthorlist}

\icmlaffiliation{nyu}{New York University}

\icmlcorrespondingauthor{Lily H. Zhang}{lily.h.zhang@nyu.edu}
\icmlkeywords{Machine Learning, ICML}

\vskip 0.3in
]

\printAffiliationsAndNotice{}  %

\begin{abstract}

\Glspl{dgm} seem a natural fit for detecting \gls{ood} inputs,
but such models have been shown to assign higher probabilities or densities to \gls{ood} images than images from the training distribution. 
In this work, we explain why this behavior should be attributed to model misestimation.
We first prove that no method can guarantee performance beyond random chance without assumptions on which out-distributions are relevant.
We then interrogate the \emph{typical set hypothesis}, the claim 
that relevant out-distributions can lie in high likelihood regions of the data distribution, and that \gls{ood} detection 
should be defined based on the data distribution's typical set.
We highlight the consequences implied by assuming support overlap between in- and out-distributions, as well as the arbitrariness of the typical set for \gls{ood} detection.
Our results suggest that estimation error is a more plausible explanation than the misalignment between likelihood-based \gls{ood} detection and out-distributions of interest, and we illustrate how even minimal estimation error can lead to \gls{ood} detection failures, yielding implications for future work in
deep generative modeling and \gls{ood} detection.

\end{abstract}

\section{Introduction}
\glsresetall Predictive models have little guarantee in performance on inputs that differ from the training distribution. Thus, detecting such \gls{ood} inputs is an important step towards safe and reliable machine learning \citep{Amodei2016ConcretePI}.
\Gls{ood} detection has been formalized as the task of identifying points with low likelihood\footnote{As is common in the literature, we will use ``likelihood'' to refer to the probability or density of a sample under a distribution, even though the statistical definition of the term refers to a function of the parameters given fixed data.}
under the training distribution, estimated via a model \cite{Bishop}.
\Glspl{dgm}  
estimate complex distributions from often high-dimensional inputs
and produce high-quality simulations \cite{Salimans2017PixelCNNIT, Chen2018PixelSNAILAI, Kingma2018GlowGF}.
However, explicit likelihood \glspl{dgm} (e.g. autoregressive models, normalizing flows) have been shown to 
assign higher likelihoods to unrelated inputs than even those from the training distribution.
For instance, a model trained on Fashion-\acrshort{mnist}, an image dataset 
of clothing items, assigns higher likelihoods to \acrshort{mnist} images. The same is true for the training distribution (or in-distribution) \acrshort{cifar-10}, a dataset of animals and vehicles, and the \gls{ood} distribution (or out-distribution) \acrshort{svhn}, a dataset of house numbers.
For such dataset pairs, \gls{ood} detection based on explicit likelihood 
\glspl{dgm} performs worse than random chance
\cite{Nalisnick2019DoDG, Hendrycks2019DeepAD}.

This observation has motivated many alternative methods for \gls{ood} detection which employ the same \glspl{dgm} but modify how they are used \cite{Ren2019LikelihoodRF, Serr2020InputCA, Schirrmeister2020UnderstandingAD, Choi2018WAICBW,DeepMind2019DetectingOI, Wang2020FurtherAO, Morningstar2020DensityOS}.
While these methods have been successful in empirical benchmarks, we 
prove that all methods are powerless against some set of out-distributions (\Cref{sec:prop1}). 
This result applies to any detection method, regardless of whether \glspl{dgm} are involved, and highlights the need to specify the out-distributions of interest for the task.

Some works have suggested that the failure of deep generative models to assign low likelihoods to \gls{ood} points is not a model failure; 
rather, out-distributions of interest can lie in high likelihood regions of the data distribution. 
To explain why points with high likelihood are never observed among samples from the data distribution, these works
mention that points assigned high density or probability under the training distribution may lie within regions of small overall probability.
A method that identifies low likelihood points as \gls{ood} will fail to detect such out-distributions, so existing
works suggest instead to flag as \gls{ood} any point which falls outside a distribution's typical set, a set that contains the majority of the probability mass of a distribution but not necessarily the highest density or probability points \cite{Choi2018WAICBW, DeepMind2019DetectingOI, Wang2020FurtherAO, Morningstar2020DensityOS}. 
This \emph{typical set hypothesis}---the idea that relevant out-distributions are determined based on the typical set of a distribution---assumes that \gls{ood} regions can lie in the support of the data distribution.

In this work, we highlight problems with the typical set hypothesis (\Cref{sec:typical_set}).
First, the hypothesis assumes that relevant out-distributions (e.g. \acrshort{svhn}) can overlap in support with the data distribution (e.g. \acrshort{cifar-10}). However, when the in- and out-distribution overlap, there is an irresolvable upper bound on \gls{ood} detection performance (\Cref{sec:single}), and
even a perfect model of the in-distribution can yield worse \gls{ood} detection than a misestimated one (\Cref{sec:decomp}). 
A preference for the typical set over other similar sets is also arbitrary for \gls{ood} detection (\Cref{sec:typical_arbitrary}).
These results highlight the implausibility of the typical set hypothesis and its support overlap assumption.

In our experiments, we offer empirical demonstrations of the analyses presented.
First, given an \gls{ood} detection method and a specific in-distribution, we provide examples of out-distributions that the method fails to distinguish from the in-distribution
(\Cref{sec:tsexp}). Then, we showcase an instance where a partially-trained \gls{dgm} yields better \gls{ood} detection than the true distribution of the data when supports overlap between the in- and out-distribution (\Cref{sec:decexp}).

Based on the implausible implications of the typical set hypothesis, we conclude that the high likelihoods assigned to certain \gls{ood} images are instead due to model estimation error.
First, it is reasonable to believe that existing dataset pairs have disjoint (rather than overlapping) support, as one would not expect to draw a house number from the \acrshort{cifar-10} distribution, or a digit from the Fashion-\acrshort{mnist} distribution, even given infinite samples. This implies that existing models mistakenly assign high probability or density where they should be assigning zero.
We demonstrate how even a model with good generation quality and heldout likelihood can still exhibit \gls{ood} failures (\Cref{sec:gen_vs_det}). We then discuss what this perspective of estimation error implies for \glspl{dgm} and \gls{ood} detection (\Cref{sec:important}). We illustrate how recent methods that were motivated by the typical set hypothesis may instead correct for model estimation error, and we suggest future modeling directions to improve \glspl{dgm} for \gls{ood} detection.

\section{Defining OOD Detection}
\label{sec:definition}
\Gls{ood} detection has been defined as the task of identifying ``whether a test example is from a different distribution from the training data'' \cite{Hendrycks2017ABF}. 
Here, we illustrate why it is critical to specify the out-distributions to consider. In fact, without any constraints on out-distributions, the task of \gls{ood} detection is impossible. 

In this section, we first formalize the broadest form of \gls{ood} detection as a single-sample goodness-of-fit test (\Cref{sec:ood_ht}). We then prove that no method can guarantee better than random chance performance under this task definition (\Cref{sec:prop1}). We conclude that any formal analysis of an \gls{ood} detection method must take into account the out-distributions which define the task.

\subsection{OOD Detection as Goodness-of-fit Testing}
\label{sec:ood_ht}

In its unconstrained form, \gls{ood} detection can be formalized
as a single-sample hypothesis test \cite{DeepMind2019DetectingOI, Serr2020InputCA, Wang2020FurtherAO}; given a sample $\mbx$, the test decides whether to reject the null hypothesis that a sample was drawn from the data distribution $P$, in favor of an alternative hypothesis that the sample came from a distribution other than $P$:
\begin{align*}
H_0&: \mbx \sim P \\
H_A&: \mbx \sim Q \in \mathcal{Q}, P \not\in \mathcal{Q}.
\end{align*}
The decision to reject or not reject the null hypothesis (i.e. mark a sample \gls{ood}) is based on the value of a predetermined test statistic $\phi$, which can be any arbitrary function of a single sample $\mbx$. In our analysis, we focus on test statistics which directly utilize knowledge of the input distribution $P$ or an estimate of it via a deep generative model $P_\theta$. 
$P_\mbtheta$ can either be a continuous distribution, as is the case for normalizing flows \cite{Dinh2015NICENI,Dinh2017DensityEU,Kingma2018GlowGF}, or a discrete distribution, as is the case for existing autoregressive models \cite{Salimans2017PixelCNNIT,Oord2016ConditionalIG, Oord2016PixelRN, Chen2018PixelSNAILAI}. 
An example of a test statistic is $\phi = \log p_\theta$, where
$p_\theta(\mbx)$ denotes either the probability of an observation for a discrete distribution or the density of an observation with respect to the Lebesgue measure for a continuous distribution. This test statistic is often accompanied by the rejection rule $\phi(\mbx) < k$, i.e. reject as \gls{ood} points where $\log p(\mbx)$ is low.
We discuss various choices of test statistics, including those which do and do not use an estimate of $P$, in the related work in \Cref{sec:related}. 

In order to determine whether an input should be processed through a given classifier,
\gls{ood} tests must make decisions on a single sample at a time. This stands in contrast with most goodness-of-fit testing setups that make a decision based on a collection of samples. We discuss the challenges of this single-sample formulation in \Cref{sec:prop1}.
The quality of a test is measured by its ability to correctly detect \gls{ood} samples without flagging in-distribution samples as \gls{ood}.
The proportion of \gls{ood} samples detected, known as the true positive rate or power of a test, can be plotted as a function of the proportion of in-distribution samples incorrectly rejected, known as the false positive rate or size of a test. This is equivalent to a \gls{roc} curve, and the \gls{auc} is the area under the power vs. size curve.

Using this general formulation of \gls{ood} detection, we 
can now interrogate \gls{ood} detection methods more broadly by abstracting away the choice of test statistic $\phi$.

\subsection{OOD Detection as a Single-Sample Distributional Test is Impossible}
\label{sec:prop1}
\gls{ood} detection defined as a single-sample goodness-of-fit test is a challenging classification task given that the out-distributions are unknown.
To remove the effect of misestimation, we consider test statistics which can use knowledge of the true in-distribution $P$ via its density or probability function, denoted $\phi_p: \mathcal{X} \rightarrow \mathbb{R}$. We now present
an impossibility result: no test can do well against all alternatives.

\begin{proposition}
\label{prop:prop1}
Let $P$ be the distribution under the null hypothesis $H_0$. Let $\mu$ be the measure associated with the distribution of test statistic $\phi_p(\mbx)$ under the null. Then, assuming the conditional $\mbx \g \phi_p(\mbx)$ is not degenerate on a $\mu$-non-measure zero set, there exists a set of alternative distributions $Q \in \mathcal{Q}$ where $Q \neq_d P$ and the test has power equal to the false positive rate. In other words, the test does no better than random guessing.
\end{proposition}

\begin{proof}
See \Cref{sec: prop1_proof}. The proof sketch is as follows: First we construct distributions $Q \in \mathcal{Q}$ for which the distribution of $\phi_p(\mbx)$ is the same but the distribution of $\mbx|\phi_p(\mbx)$
differs when $\mbx \sim P$ and $\mbx \sim Q$
for all $\phi_p(\mbx)$ in a non-measure-zero set $\Phi$. This implies $q(\mbx) \neq_d p(\mbx)$.
We show that the power of the test for any rejection rule for such a pair $P, Q$ is equal to the false positive rate for all false positive rates, which is equivalent to random guessing. 
\end{proof}

\Cref{prop:prop1} demonstrates that no test statistic can be useful for all possible out-distributions. 
In the context of single-sample distributional testing, 
all proposed test statistics trade off power against different out-distributions. 
This means that, without additional assumptions on the family of alternative hypotheses for \gls{ood} detection, no test statistic can be uniformly better across out-distributions than another.
To build intuition behind the proposition, imagine that $\mathcal{X}$ is the space of $d$-dimensional reals $\mathbb{R}^d$ and the in-distribution has a density with respect to the 
$d$-dimensional Lebesgue measure. 
The test statistic is a function that maps from $\mathbb{R}^d \to \mathbb{R}$; thus, the statistic is a one-dimensional projection of the distribution. In the same way that not all differences in two multivariate distributions can be assessed by looking at a single marginal, not all the differences between $P$ and $Q$ can be assessed by looking at their projections on the test statistic. This result is focused on the single-sample formulation of \gls{ood} detection and holds even for test statistics which are consistent in power asymptotically. 

\paragraph{An Example: Using $\log p$ as a test statistic.}
When the test statistic is the log probability or density, 
the set of alternative distributions $Q \in \mathcal{Q}$ that cannot be distinguished from $P$ are those which yield the same distribution of log probabilities or densities under $P$. These are distributions which collapse any of the level sets of $P$. As an example in the discrete case, imagine a countable sample space and a distribution $P$ where $c$ of the elements are given the same probability.
Any distribution $Q$ which moves the total probability of the $c$ elements in $P$ to any subset of these elements will share the same distribution of probability under $P$. 
The analogue for continuous distributions $\mathbb{R}^d$ is collapsing level sets of dimension $\mathbb{R}^{d-1}$.
We illustrate the phenomenon with an example in the continuous case in \Cref{sec:tsexp}.

\Cref{prop:prop1} emphasizes the need to specify the family of relevant out-distributions for \gls{ood} detection. For instance, likelihood ratio test statistics \cite{Ren2019LikelihoodRF, Serr2020InputCA, Schirrmeister2020UnderstandingAD} are optimal when the alternative hypothesis is correctly specified, but like all test statistics, they trade off power in some other alternative; therefore, comparing different likelihood ratios (and test statistics in general) is only useful when the family of out-distributions is formalized and standardized. 

Like any test statistic, $\phi = \log p$ works well for some out-distributions (those whose samples have zero or low likelihood under the data distribution) but poorly for others (those whose samples have high likelihood under the data distribution).
Whether the latter such out-distributions are relevant for \gls{ood} detection is a central question underlying our analysis of the typical set hypothesis.

\section{The Implausibility of the Typical Set Hypothesis}
\label{sec:ts}
\citet{Nalisnick2019DoDG, Hendrycks2019DeepAD} observed that \glspl{dgm} trained on \acrshort{cifar-10} samples assign higher likelihoods to \acrshort{svhn} images, and \glspl{dgm} trained on Fashion\acrshort{mnist} samples assign higher likelihoods to \acrshort{mnist} images. 
The explanation for these observations can either be A. such \gls{ood} samples do have high likelihoods under the \emph{data} distribution, or B. these \gls{ood} samples only have high likelihoods under the \emph{model} distribution due to estimation error.
The typical set hypothesis argues for the former,
that out-distributions can lie in high probability or density regions of the data distribution.
A test based on the $\log p$ test statistic and $\phi(\mbx) < k$ rejection rule lacks power, even under the perfect model, to detect out-distributions whose samples have high likelihood under the data distribution, and the typical set hypothesis assumes that \acrshort{svhn} is such an out-distribution relative to \acrshort{cifar-10}, as is \acrshort{mnist} to Fashion-\acrshort{mnist}.
In this section, we detail the typical set hypothesis
(\Cref{sec:typical_set}),
reveal consequences which fall from its assumptions (\Cref{sec:single}, \Cref{sec:decomp}),
and discuss its relevance for \gls{ood} detection (\Cref{sec:typical_arbitrary}).

\subsection{The Typical Set Hypothesis}
\label{sec:typical_set}
The typical set hypothesis posits that 1. out-distributions of interest can lie in regions of high likelihood but small overall probability under the data distribution, and 2. to detect such distributions, \gls{ood} detection should take into account the data distribution's typical set\footnote{\citet{DeepMind2019DetectingOI} discuss the \emph{model's} typical set rather than the data distribution's but do not mention model estimation error. Subsequent works have interpreted their message in the context of the data distribution's typical set 
\cite{Morningstar2020DensityOS}.} \cite{Choi2018WAICBW, Nalisnick2019DoDG, Wang2020FurtherAO, Morningstar2020DensityOS}.
Tests that reject
low likelihood points will perform worse than random chance on out-distributions whose samples have high in-distribution likelihoods, but tests 
that consider
the data distribution's typical set can have power over such out-distributions.
Quoting \citet{Wang2020FurtherAO}:
\begin{quote}
Samples
from a high-dimensional distribution will often fall on a typical set with high probability, but the
typical set itself does not necessarily have the highest probability density at any given point. Per this
line of reasoning, to determine if a test sample is an outlier, we should check if it falls on the typical
set of the inlier distribution rather than merely examining its likelihood under a given deep generative model.
\end{quote}

Given a distribution $P$, the typical set $A_\epsilon^{(n)}$ is the set of $n$-length sequences $(x_{i1}, ..., x_{in}), x_{ij} \stackrel{\text{i.i.d.}}{\sim} P$
whose empirical entropy is close to the entropy of $P$, i.e. $H(P) = \minus\mathbb{E}_{x_{ij} \sim P}[\log p(x_{ij})]$, within a neighborhood determined via the constant $\epsilon$ \cite{Cover1991ElementsOI}:
\begin{equation} \label{eq:typical}
  H(P) - \epsilon \leq -\frac{1}{n} \sum_{j=1}^n \log p(x_{ij}) \leq H(P) + \epsilon.  
\end{equation}
The typical set can be viewed as a set of elements from the sample space of the product measure $P^n = P \times P  \times ... \times P$ ($n$ copies). 
For a sufficiently large $n$, i.e. a sufficiently high-dimensional distribution $P^n$, the typical set is small relative to the total number of possible elements in $P^n$, yet the probability of the set under $P^n$ is close to one. 
The idea underlying the typical set hypothesis is the following: 
If nearly all of the total probability mass of a distribution is concentrated in a small set, then it is unlikely that a sample generated from the distribution will fall outside of this set. For instance, samples from an out-distribution such as \acrshort{svhn} could have high likelihood in the \acrshort{cifar-10} distribution but fall outside its typical set, which would explain why \acrshort{svhn} samples are not seen in the finite \acrshort{cifar-10} dataset. 

Tests based on the typical set have power to detect out-distributions concentrated in high likelihood regions of the data distribution which have small overall probability, but the assumption that an out-distribution like \acrshort{svhn} lies within the support of a data distribution like \acrshort{cifar-10} yields questionable consequences, illustrated in the next two sections.

\subsection{No Method Can Guarantee Perfect Detection When Supports Overlap}
\label{sec:single}
In order to explain why \glspl{dgm} trained on \acrshort{cifar-10} or Fashion-\acrshort{mnist} place high likelihood on \acrshort{svhn} or \acrshort{mnist} samples respectively, the typical set hypothesis must assume that the out-distributions overlap in support with the in-distributions. 
However,
the probability of classification error is non-zero when the support of a given out-distribution $Q$ overlaps with that of the in-distribution $P$.
Therefore, even with exact knowledge of the in-distribution, no method can achieve perfect detection against out-distributions which overlap in support with the in-distribution.

\begin{proposition}
\label{prop:prop2}
Let $P$ and $Q$ have overlapping support: $\textrm{Pr}_Q(\mbx \in 
\text{supp}(p(\mbx))) > 0$. Then, any test has non-zero probability of error.
\end{proposition}
\begin{proof}
Assume there exists a rejection rule $\phi_p(\mbx) \not\in \Phi$ that perfectly separates samples from $P$ and $Q$ i.e.
\begin{align*}
	\textrm{Pr}_Q(\phi_p(\mbx) \in \Phi) = 0 \textrm{, and } \textrm{Pr}_P(\phi_p(\mbx) \in \Phi) = 1.
\end{align*}
The above condition requires $\Phi$ to encompass all values in $\{\phi(\mbx) | \mbx \in \text{supp}(p(\mbx))\}$ and none in $\{\phi(\mbx) | \mbx \in \text{supp}(q(\mbx))\}$. However, since $\textrm{Pr}_Q(\mbx \in 
\text{supp}(p(\mbx))) > 0$, $\textrm{Pr}_Q(\phi_p(\mbx) \in 
\text{supp}(p(\phi_p(\mbx)))) > 0$. By contradiction, there exists no subset $\Phi$ that perfectly separates $P$ and $Q$.
\end{proof}

\Cref{prop:prop2} states that if the supports of two distributions (e.g. \acrshort{svhn} and \acrshort{cifar-10}) overlap, then no solution can guarantee perfect discrimination between single samples from these two distributions. This relates to the bound on performance given by the Bayes optimal classifier: even the optimal classifier has non-zero error when the covariate distributions from two classes overlap.

\subsection{A Wrong Model Can Perform Better Than a Perfect One When Supports Overlap}
\label{sec:decomp}
An additional consequence of including support overlap cases in \gls{ood} detection is that for a given out-distribution $Q$, a perfect model can perform worse than a misestimated one.

Define $\phi_p$ using the data distribution $P$ (e.g. $\phi_p = \log p$)
such that the rejection rule is of the form $\phi_p(\mbx) < k$.
(We can recast existing test statistic and rejection rule pairings to follow this form, even if the original pairing does not use rejection rule $\phi_p(\mbx) < k$.  See \Cref{sec:rej} for details.) 
 We can write the \gls{auc} of an \gls{ood} detection procedure using $\phi_p$ as $\textrm{Pr}(\phi_p(\mbx) > \phi_p(\textbf{y}))$ for $\mbx \sim P, \textbf{y} \sim Q$.  Perfect discrimination is achieved when $\textrm{Pr}(\phi_p(\mbx) > \phi_p(\textbf{y})) = 1$.

We now show that it is possible for \gls{ood} detection based on a misestimated model to perform better than detection using the true in-distribution when the supports of $P$ and $Q$ overlap. Let $\phi_p = p$, and let $Q$ have support over the entire sample space $\mathcal{X}$. We can construct a $P_\theta$ proportional to the likelihood ratio of $P$ and $Q$:
\[p_\theta(\mbx) = \frac{1}{C}\frac{p(\mbx)}{q(\mbx)}, C = \int_{\mathcal{X}} \frac{p(\mbx)}{q(\mbx)} d\mbx, \]
assuming integrability. Then, $\phi_{p_\theta}$ is proportional to the likelihood ratio, and by the Neyman-Pearson Lemma \cite{Neyman1933OnTP}, a likelihood ratio test statistic is uniformly most powerful for a simple hypothesis (i.e. yields the highest power against any single alternative hypothesis $Q$ for a specified test size). Since the ratio is most powerful at every false positive rate specified, \gls{ood} detection via $\phi_{p_\theta}$ achieves the maximal \gls{auc} possible for a given pair $P, Q$. Since a uniformly most powerful decision rule is unique up to sets of measure zero, given $P_\theta \neq P$, \gls{ood} detection using $P_\theta$ is strictly better than detection using $P$. In summary, 
\[\textrm{Pr}(\phi_{p_\theta}(\mbx) > \phi_{p_\theta}(\textbf{y})) > \textrm{Pr}(\phi_p(\mbx) > \phi_p(\textbf{y})).\]

The same idea applies even when $Q$ does not place strictly positive density or probability across the sample space $\mathcal{X}$ or when the test statistic is a function other than $\phi_p = \log p$: A model $P_\theta$ which makes values of $\phi_{p_\theta}(\mbx), \mbx \sim P$ higher relative to $\phi_{p_\theta}(\mby), \mby \sim Q$ will improve \gls{ood} detection. 
We illustrate this phenomenon with an empirical example in \Cref{sec:decexp}, comparing the \gls{ood} performance for a specific $P, Q$ pair when the test statistic utilizes the true distribution $P$ versus a (poor) estimate of it.

Note that this result does not apply when distributions $P$ and $Q$ have disjoint support and the test statistic used is $\phi_p = p$.
Concretely, $\phi_p(\mby) = 0$ for all $\mby \sim Q$ and $\phi_p(\mbx) > 0$ for all $\mbx \sim P$, which implies that $\textrm{Pr}(\phi_p(\mbx) > \phi_p(\textbf{y})) = 1$; in this setting, likelihood-based \gls{ood} detection using a perfect model yields optimal performance.

\subsection{OOD Detection based on the Typical Set is Arbitrary}
\label{sec:typical_arbitrary}
The previous two sections (\Cref{sec:single} and \Cref{sec:decomp}) highlight two consequences that result from the assumption in the typical set hypothesis that out-distributions of interest (e.g. \acrshort{svhn}, \acrshort{mnist}) overlap in support with the in-distribution (e.g. \acrshort{cifar-10}, Fashion\acrshort{mnist}).
Beyond the issues resulting from the support overlap assumption, the typical set hypothesis relies on the idea that the typical set is the preferred subset to demarcate what is in- and out-of-distribution. Here, we question this idea, explaining why the properties of the typical set do not suffice to explain its relevance to \gls{ood} detection.

As discussed in \Cref{sec:typical_set}, the typical set hypothesis uses the fact that the typical set can be small yet high probability to justify why points outside the set should be considered \gls{ood}.
However, there can exist other similarly small sets that also contain nearly all of the probability mass, meaning it is arbitrary to prefer the typical set based on its small size and high probability properties alone.

As an example, consider 
a high-dimensional distribution of i.i.d. Bernoullis, each with 75\% probability of success. A vector of 100\% ones, denoted $\mathbf{z}_{100}$, has the highest probability but is not part of the typical set
$\mathcal{A}_\epsilon$ which consists of sequences with close to 75\% ones and 25\% zeros.
The property $\Pr(\mbx \in \mathcal{A}_\epsilon) \approx 1$ is used to justify why it is okay to consider any points outside this set as \gls{ood}, including $\mathbf{z}_{100}$. 
However, we also can define a new set $\mathcal{A}'_\epsilon$ which substitutes in $\mathbf{z}_{100}$ in place of one of the 72/25 sequences---for instance, the sequence whose first 75\% of elements are ones and last 25\% are zero, which we will denote $\mathbf{z}_{75}$. This same-sized set also satisfies $\Pr(\mbx \in \mathcal{A}'_\epsilon) \approx 1$ and in fact has strictly greater probability than the typical set, since a sequence of ones is more likely than the particular 75/25 sequence that was removed. Under this newly constructed set, the sequence $\mathbf{z}_{75}$ would be considered \gls{ood} since it is not contained in $\mathcal{A}'_\epsilon$. Since we randomly selected one of the 75/25 sequences to replace, the decision to mark this sequence \gls{ood} based on set membership is arbitrary. Yet, it is just as arbitrary to exclude $\mathbf{z}_{100}$ from the set of in-distribution points; after all, $\Pr(\mathbf{z}_{100}) > \Pr(\mathbf{z}_{75})$.

The unique property of the typical set relative to other small-volume, high-probability sets is that its elements are close to equiprobable. However, the need for this property does not follow from the motivation for \gls{ood} detection.

In summary, there are several issues with the typical set hypothesis.
First, the hypothesis relies on the assumption that out-distributions can overlap in support with the in-distribution; this support overlap assumption implies that perfect discrimination between image datasets distibutions such as \acrshort{cifar-10}/\acrshort{svhn} and Fashion-\acrshort{mnist}/\acrshort{mnist} is impossible (\Cref{sec:single}), and that one could perform better \gls{ood} detection with a wrong model of the data distribution than the right one (\Cref{sec:decomp}). Additionally, there is no clear motivation for preferring the typical set over other small volume, high probability sets. These issues suggest the implausibility of the typical set hypothesis as the explanation and solution for existing \gls{ood} failures of \glspl{dgm}. 

Consequently, the phenomenon observed in \citet{Nalisnick2019DoDG, Hendrycks2019DeepAD} is likely a result of model estimation error, rather than a property of existing image data distributions requiring an alternative task definition. We detail what this perspective implies about existing models and provide guidance for future work in \Cref{sec:model_perspective}.

\section{Experiments}
The following experiments demonstrate the theoretical properties shown in the prequel. To build intuition around the impossibility result of \Cref{prop:prop1}, we give an example of different distributions with the same distribution of densities under $P$, meaning a test based on a likelihood statistic cannot distinguish them even with access to a perfect model of the in-distribution.
We then demonstrate an instance of a partially-trained \gls{dgm} performing better \gls{ood} detection than the actual model of the data, providing an empirical example for the analysis in \Cref{sec:decomp} that an erroneous model can perform better \gls{ood} detection than a perfect one.

\subsection{Any Test Statistic has Failure Modes When Possible Out-distributions are Unrestricted}
\label{sec:tsexp} 
\Cref{prop:prop1} states that any test statistic gives up power over certain alternative hypotheses. We show that the test statistic $\phi(\mbx) = \log p(\mbx)$ cannot detect as \gls{ood} single samples drawn from out-distributions $Q, R$ which are contained within $P$ but collapse any of its level sets. 

\begin{figure}
    \centering
    \subfigure[Samples from three different distributions]{\label{fig:1a}\includegraphics[width=40mm]{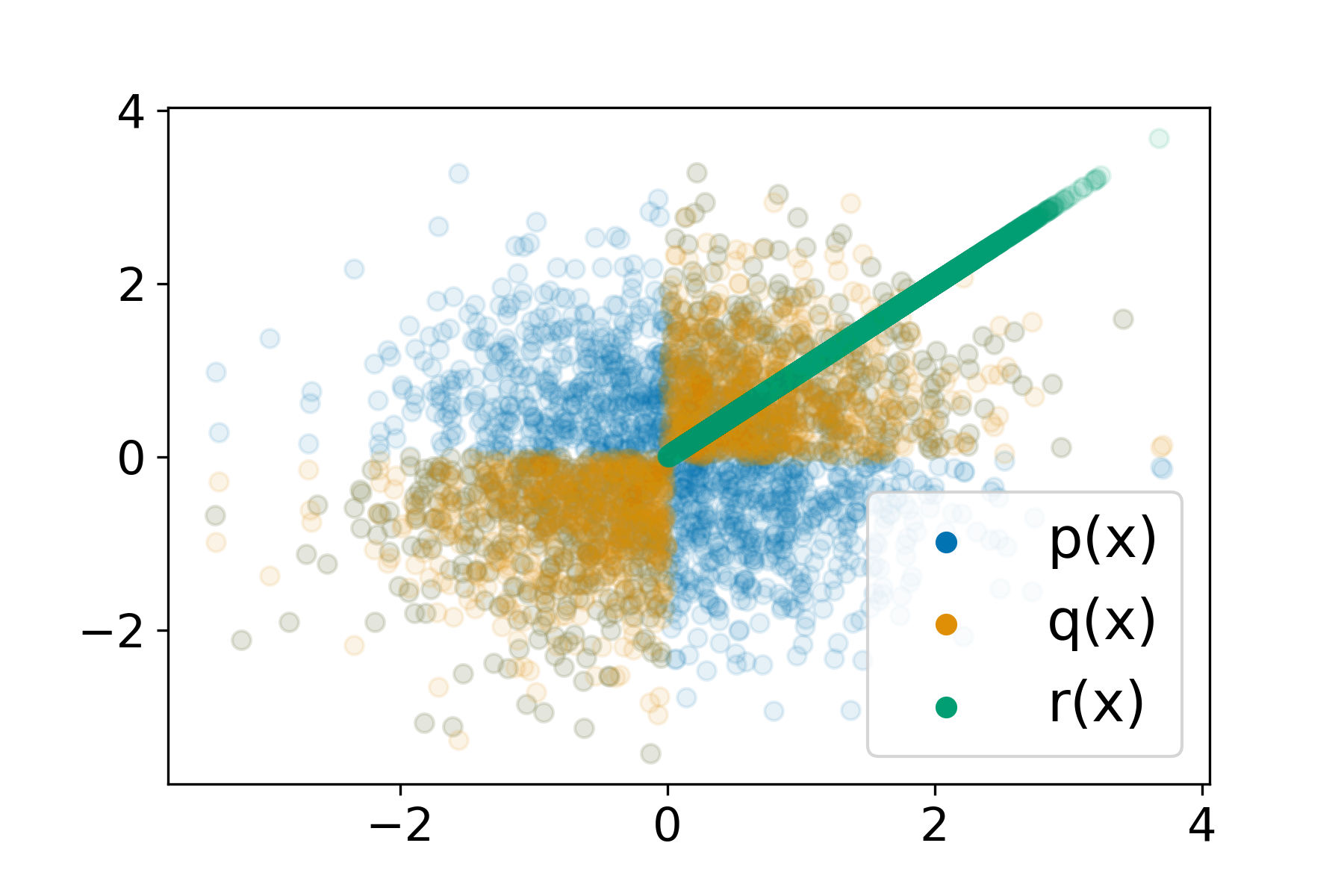}}  %
    \subfigure[Distributions of the log-density under $P$]{\label{fig:1b}\includegraphics[width=40mm]{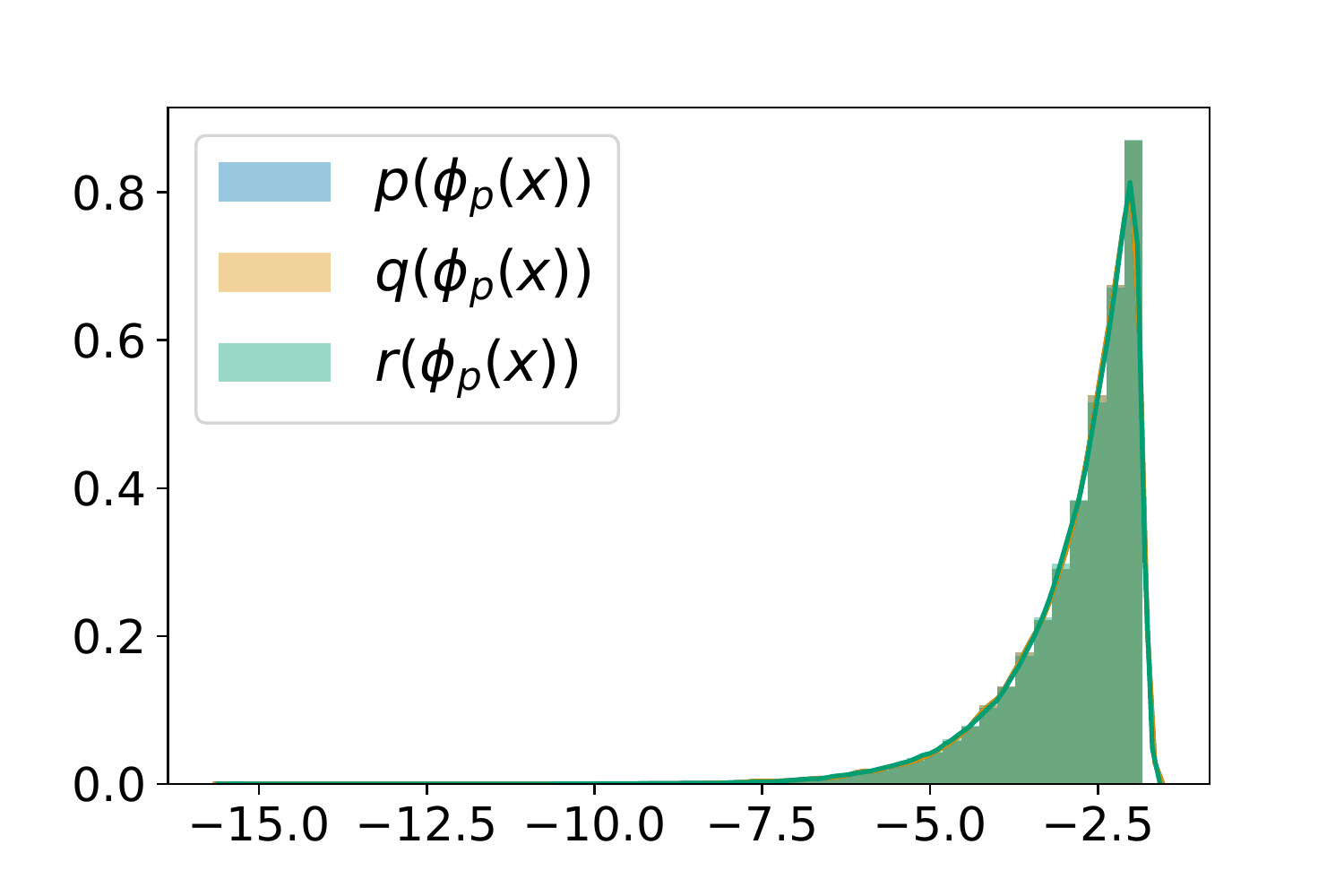}}
    \caption{For a given in-distribution $P$ and choice of test statistic, we can construct out-distributions where \gls{ood} detection will perform no better than random chance. In the above example, given test statistic $\phi_p(\mbx) = \log p(\mbx)$, we can construct out-distributions $Q, R$ (left) such that the distribution of log-density under $P$, i.e. $\phi_p(\mbx)$, is the same regardless of whether $\mbx$ comes from $P, Q$ or $R$ (right).
    }
    \label{fig:prop1}
\end{figure}

Consider an in-distribution $P$ that is bivariate Gaussian with an identity covariance matrix. There are a variety of out-distributions whose samples yield the same distribution of the test statistic $\phi(\mbx) = \log p(\mbx)$. We consider two in \Cref{fig:prop1}: $Q$, the distribution obtained by
sampling $(x_1,x_2)$ from a standard bivariate normal and then flipping the sign of $x_2$
if it is in the second or fourth quadrant, and $R$, the distribution
obtained by sampling $(x_1,x_2)$ from a standard bivariate normal and mapping it to the point $(z,z)$ where $z^2 = (x_1^2 + x_2^2)/2$ (i.e. preserving distance from origin). 
The out-distributions $Q$ and $R$ maintain the same distribution of log-densities under $P$ as the distribution $P$ since they distribute mass similarly across the upper level sets $\{\mbx: p(\mbx) > t\} \forall t$. 
We can use similar logic to construct problematic out-distributions for other test statistics which are a function of $p(\mbx)$, including the test statistic associated with the typicality test in \citet{DeepMind2019DetectingOI}. 
In fact, this test statistic, $\phi(\mbx) = \big \lvert \minus \log p(\mbx) - \hat{H}_p \big \rvert$ where $\hat{H}_p =  \minus \frac{1}{\lvert \mathcal{D}_{tr}\rvert}\sum_{\mbx \in \mathcal{D}_{tr}} \log p(\mbx)$, is no better than random chance 
whenever a log-probability test statistic is no better than chance.
To see this, note that $\hat{H}_p$ is a constant, so the resulting distributions over $\phi(\mbx)$ are simply shifted and scaled relative to the distributions over the log-likelihoods. 

\subsection{A Bad Model Can Beat a Perfect One When the Out- and In-Distributions Overlap in Support}
\label{sec:decexp}
\begin{figure}
    \centering
    \subfigure[True model, \gls{auc}=.96]{\label{fig:3a}\includegraphics[width=40mm]{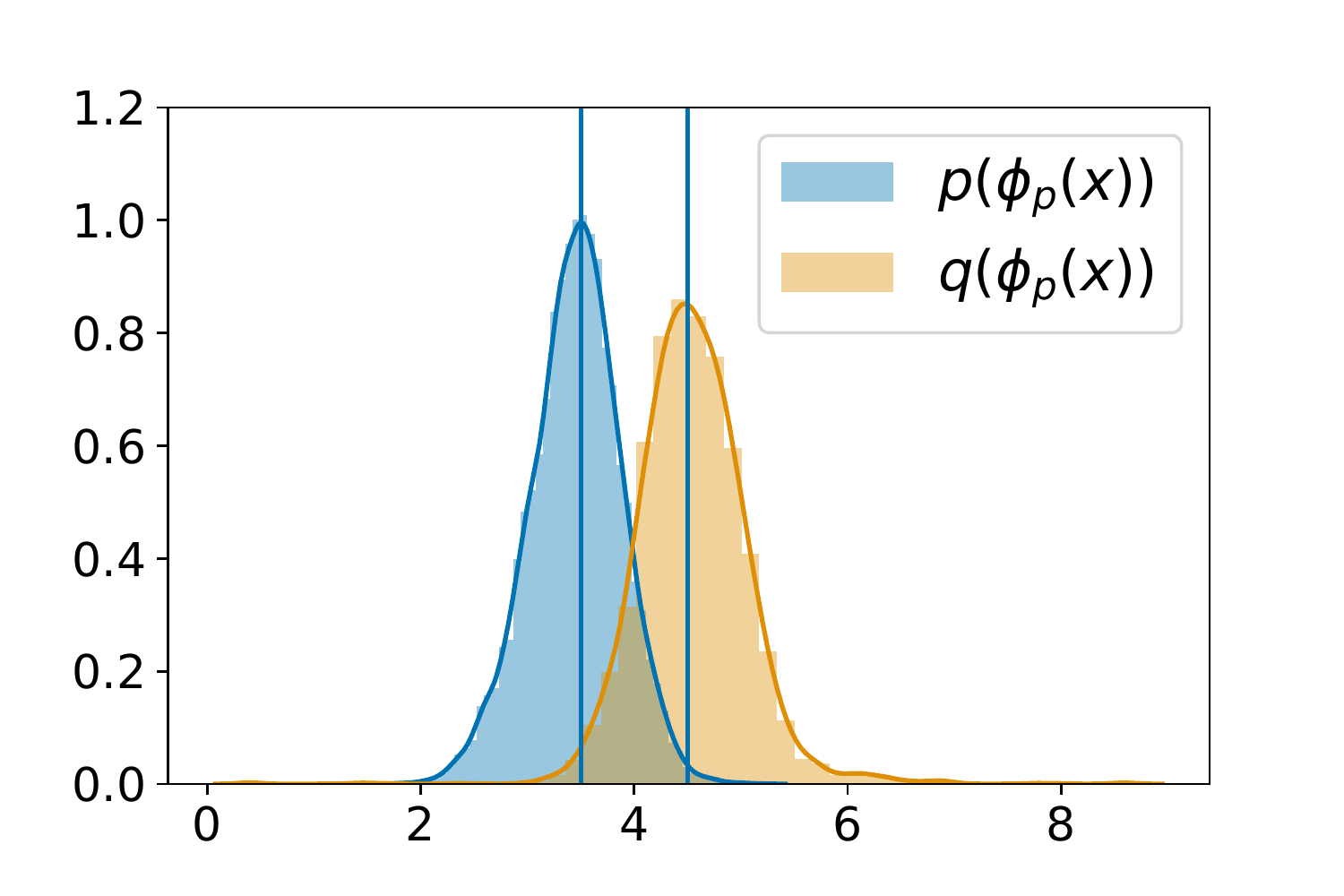}}
    \subfigure[DGM, \gls{auc}=.98]{\label{fig:3b}\includegraphics[width=40mm]{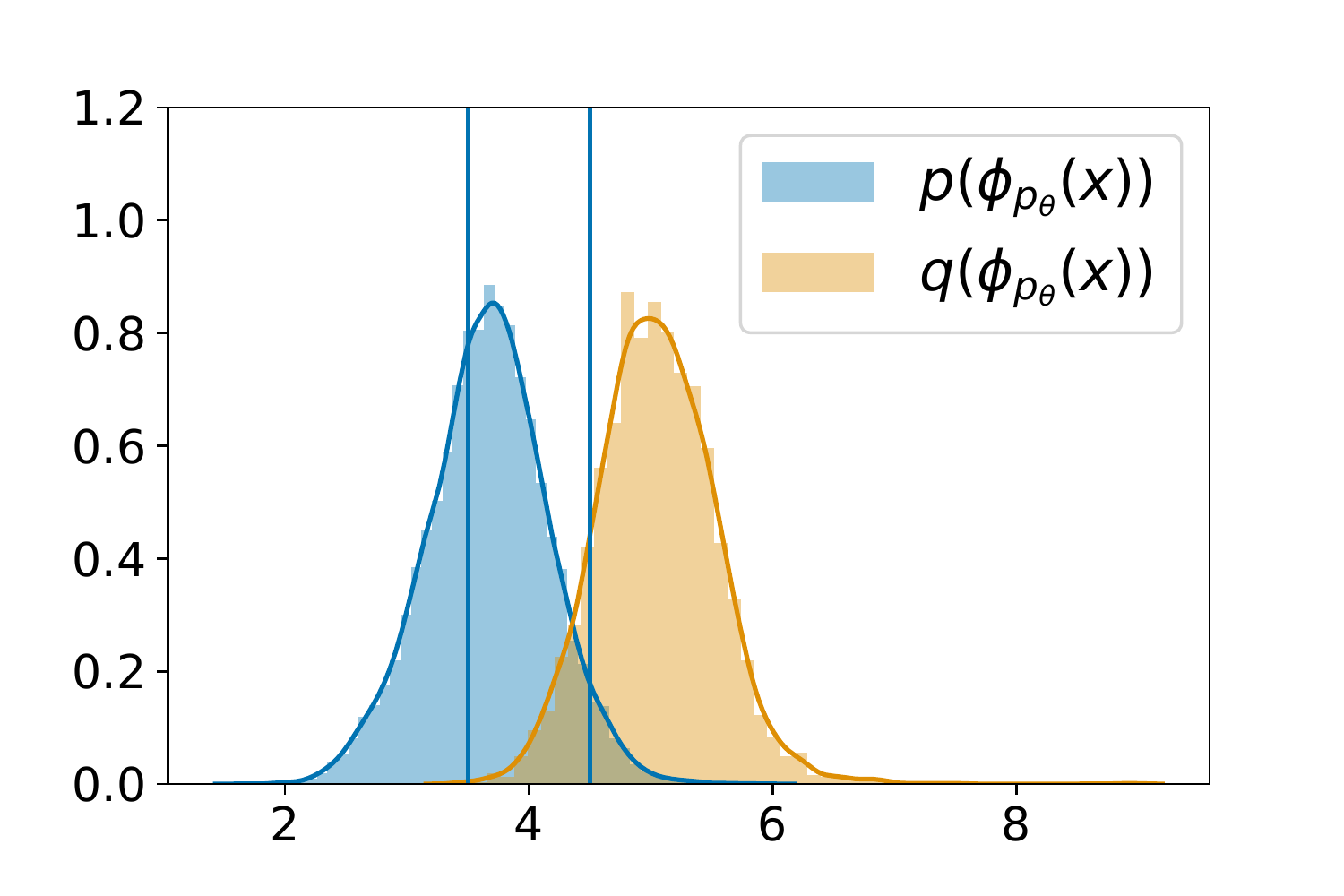}}
    \caption{A perfect model can perform worse than a misestimated one when supports of the in- and out-distributions overlap. 
    The \gls{dgm} yields better \gls{ood} detection performance than the true model because the misestimation has decreased the amount of overlap in the distributions of the test statistic bits per dimension.}
    \label{fig:decomp}
\end{figure}

We now provide an empirical example where misestimation can result in better \gls{ood} detection of a particular out-distribution when supports of the in- and out-distribution overlap.
Because we require access to the true model density, we designate a pretrained \gls{dgm} as our in-distribution $P$---specifically the \acrshort{glow} model of \citet{Kingma2018GlowGF} trained on \acrshort{cifar-10}. Next, we train a separate \acrshort{glow} model $P_\mbtheta$ on 40,000 samples from $P$. See \Cref{sec:decexp_supp} for model and training details. We then compare performance of an \gls{ood} detection method using $p$ versus $p_\mbtheta$. We choose the CelebA dataset of celebrity faces \cite{liu2015faceattributes} as our out-distribution $Q$. The in-distribution $P$ represented by the flow overlaps in support with the out-distribution $Q$, as evidenced by the fact that 
the pretrained model assigns positive densities to all CelebA images. Our partially-trained model $P_\theta$ (only 50 epochs) achieves an average
\gls{bpd} of 3.67 on the test samples from $P$, versus 3.45 for the true model (lower is better).
However, $P_\theta$ improves \gls{ood} performance relative to the true model for 
this choice of $Q$. The misestimation has increased \gls{bpd} for both the in-distribution and out-distribution samples relative to the true model; however, since the extent of the increase is higher for the out-distribution samples, the result is better separation, meaning better \gls{ood} detection under the misestimated model.
See \Cref{fig:decomp}.

\section{An Alternative Perspective}
\label{sec:model_perspective}

Rather than conclude that existing image data distributions assign high likelihoods to certain \gls{ood} images, we now consider an alternative explanation, that the phenomenon observed in \citet{Nalisnick2019DoDG, Hendrycks2019DeepAD} is a result of model estimation error.

First, 
it is reasonable to assume that the supports of dataset pairs such as \acrshort{cifar-10} and \acrshort{svhn} are disjoint: for instance, we would not expect to draw a house number from the true \acrshort{cifar-10} distribution even given infinite samples. 
This assumption is untestable, but if it does hold, then
existing \glspl{dgm} are mistakenly assigning high probability or density in places where they should be assigning none.

Such misestimation may seem surprising given that a \gls{dgm} trained on \acrshort{cifar-10} never seems to generate \acrshort{svhn} images, as previous works have noted.
We can understand this as 
poor estimation in small-volume regions of the sample space with negligible total probability mass. In fact, even good generators can be very wrong 
in this regard. 
From this perspective, training a \gls{dgm} for \gls{ood} detection can require accurate estimation in regions which are unimportant for good generation. We demonstrate this with an example.

\subsection{A Good Generator Can Still Exhibit OOD Detection Failures}
\label{sec:gen_vs_det}
A model $P_\theta$ can output samples from the true data distribution $P$ with probability close to 1 (i.e. good generation) yet still assign higher probability or density to certain out-of-support samples (i.e. bad \gls{ood} detection). As an illustration, consider a finite, discrete sample space where distributions $P$ and $Q$ have disjoint support. If the size of the support of $P$ 
is much greater than that of the support of $Q$, i.e. $\lvert \text{supp}(P) \rvert >>> \lvert \text{supp}(Q) \rvert$, then a model $P_\theta$ could place higher probability mass on each element of $Q$ than any element in $P$ yet assign negligible probability denoted by $\epsilon$ in total to the elements in $\text{supp}(Q)$, i.e.
\begin{align}
\nonumber &\textrm{Pr}_{p_\theta}(\text{supp}(P)) = 1 - \epsilon, \quad \textrm{Pr}_{p_\theta}( \text{supp}(Q)) = \epsilon,\\
&\textrm{Pr}_{p_\theta}(\mbx) < \textrm{Pr}_{p_\theta}(\mathbf{y}), \,\, \forall \, \mbx \in \text{supp}(P), \mathbf{y} \in \text{supp}(Q)
\label{eq:gen_vs_det}
\end{align}

As an example, let $P$ be a distribution which assigns the same probability to each element in its support, i.e. $\textrm{Pr}_{p}(\mbx) = 1/\lvert \text{supp}(P) \rvert$. 
Consider $P_\theta$ which moves $\epsilon / \lvert \text{supp}(P) \rvert$ probability away from each element in $\text{supp}(P)$ and splits it equally across all $\mathbf{y} \in \text{supp}(Q)$, i.e. $\textrm{Pr}_{p_\theta}(\mathbf{y}) = \epsilon /\lvert \text{supp}(Q) \rvert$. 
To meet the criteria of \Cref{eq:gen_vs_det},
\begin{align*}
    \epsilon > \frac{\lvert \text{supp}(Q) \rvert} {\lvert \text{supp}(P) \rvert + \lvert \text{supp}(Q) \rvert}.
\end{align*}
When $\lvert \text{supp}(P) \rvert$ is much larger than $\lvert \text{supp}(Q) \rvert$,
the $\epsilon$ needed for $P_\theta$ to assign higher probabilities to points
in the support of $Q$ is small.
Such a model would also have very good in-distribution probabilities, smaller than those of $P$ only by an $\epsilon/\lvert \text{supp}(P) \rvert$ amount. In other words, even small differences in held-out log probabilities can matter for \gls{ood} detection of small-volume out-distributions. For instance, using $\epsilon = \nicefrac{\lvert \text{supp}(Q) \rvert}{ \lvert \text{supp}(P) \rvert}$, if $\lvert \text{supp}(P)\rvert = 10^6$ and $\lvert \text{supp}(Q)\rvert = 10^4$, then $\epsilon = 10^{\minus 2}$ and $\textrm{Pr}_{p}(\mathbf{x}) = 10^{\minus 6}$ while $\textrm{Pr}_{p_\theta}(\mathbf{x}) = 10^{\minus 6} - 10^{\minus 8}$. The resulting negative log-likelihoods are -13.8155 and -13.8255, respectively. When $\lvert \text{supp}(Q) \rvert$ is even smaller relative to $\lvert \text{supp}(P) \rvert$, the differences can become even smaller. See \Cref{tab:supp} for examples.

\begin{table}[h] 
\centering
\caption{ \label{tab:supp} A model $P_\theta$ can place higher probabilities on an out-distribution $Q$ even with in-distribution negative log-likelihoods (NLL) close to the true model $P$ (Oracle). All calculations are based on a uniform $P$ with $\lvert\text{supp}(P)\rvert = 10^6$ and a uniform transfer of $\epsilon =  \nicefrac{\lvert \text{supp}(Q)9 \rvert}{\lvert \text{supp}(P) \rvert}$ from $P$ to be divided evenly across $\lvert \text{supp}(Q) \rvert$ of different size ($10^4$, $10^3$, $10^2$).}
\begin{tabular}{c?ccc}
\toprule
& \multicolumn{3}{c}{$\lvert \text{supp}(Q) \rvert$} \\
\hline
Oracle & $10^4$ & $10^3$ & $10^2$ \\
-13.8155 & -13.8255 & -13.8165 & -13.8156 \\
\bottomrule
\end{tabular}
\vskip -.1in
\end{table}

While the above example considers discrete distributions, the same idea holds for continuous probability measures with bounded support.\footnote{Let densities $p$ and $q$ be defined with respect to probability measures $P$ and $Q$ and base Lebesgue measure $\mu$. Consider 
$\mu(\text{supp}(P)) >>> \mu(\text{supp}(Q))$.}
We can construct examples similar to the one above by transferring a small amount of mass, originally spread over a large volume within $\text{supp}(P)$, to a region of small volume outside of $\text{supp}(P)$, e.g. $\text{supp}(Q)$.
\citet{Choi2018WAICBW, DeepMind2019DetectingOI, Wang2020FurtherAO}
have made similar points
to explain how
there can exist
elements with
high density or probability that are almost never generated.
While these works have sought to characterize the true in-distribution $P$, 
we describe this phenomenon strictly in 
terms of model misestimation $P_\theta \neq P$.

This scenario can apply to existing \gls{ood} failures if the volume of the support of the out-distribution  is small in comparison to that of the in-distribution. While it is difficult to determine whether this difference in volumes exists in these image distributions, there is reason to believe it might. For one, there are many more pixel combinations which generate a textured pattern than a smooth one. As an extreme example, consider a distribution over solid-color images versus one of random noise images. The former consists of elements where the image is perfectly predictable from the first pixel, and the size of the support is bounded by the number of possible first-value pixels. The latter has much larger support.
Likewise, it is plausible that image distributions containing varying textures (e.g. \acrshort{cifar-10}, Fashion-\acrshort{mnist}) have larger support than distributions with smooth textures (e.g. \acrshort{svhn}, \acrshort{mnist}). This suggests that volume differences can exist in real image distributions.
The fact that a good generator can still experience \gls{ood} detection failures suggests that good generation (and a high held-out likelihood) is not sufficient for good \gls{ood} detection.
Poor model likelihoods over relatively small-volume regions may be a concern for \gls{ood} detection. 

\subsection{Improving OOD Detection with DGMs}
\label{sec:important}
Given this perspective that model estimation error is the problem, we list several ways to improve \gls{ood} detection. We first discuss how alternative test statistics, including existing ones, can correct for known errors. We then turn to potential future directions to address model bias. 

First, given knowledge of a particular model bias, we can construct alternative test statistics to ameliorate the bias for the application of \gls{ood} detection. For instance, consider the issue of \glspl{dgm} placing high probability or density where they should place zero. Assuming the distribution of $\log p_\theta(\mbx)$ is the same for a test set drawn from the same distribution as the training set, \gls{ood} images which are assigned higher likelihoods than training images will be further away from the average training likelihood than the in-distribution test images will be. Consequently, a test statistic which considers distance to the average training likelihood---the statistic proposed by \citet{DeepMind2019DetectingOI}---will perform better \gls{ood} detection than $\phi = \log p_\theta$. We can improve this further by fitting a non-parametric density estimator over the probabilities assigned to the training images and rejecting when a particular likelihood value has not yet been seen or is rare---this \gls{ood} procedure is a simplified version of the density of states estimator proposed in \citet{Morningstar2020DensityOS}, who consider a several statistics jointly, not just the likelihood under the model.
While test statistics such as those in \citet{DeepMind2019DetectingOI, Morningstar2020DensityOS} were introduced to approximate an alternative definition of \gls{ood} based on the typical set, a more plausible viewpoint is that these test statistics correct for estimation error off the support of the in-distribution.

Alternative test statistics can help in certain cases, but they are not guaranteed to improve detection results over $\log p$ across out-distributions, as shown empirically \cite{Morningstar2020DensityOS}. An alternative fix involves improving the models themselves. To avoid the issues observed in \citet{Nalisnick2019DoDG, Hendrycks2019DeepAD}, it is important that models can sufficiently push down probability or density outside of the support of the in-distribution.
To do so may require different modeling preferences than the ones developed with other applications, like generation, of \glspl{dgm} in mind. As an example, certain inductive biases such as convolutional layers which benefit image modeling in general may make \gls{ood} detection between image datasets more difficult; \citet{Schirrmeister2020UnderstandingAD} found that replacing the convolutional layers in a \acrshort{glow} model with fully connected layers improved \gls{ood} detection of problematic image dataset pairs, even though it resulted in worse likelihoods. 

The choice of objective may matter as well. \Gls{mle} minimizes $\text{KL}(p||p_\theta)$ which does not allow $p_\theta$ to be zero anywhere $p$ is non-zero \cite{jerfel2021variational}.
This means maximum likelihood favors overdispersed solutions
in the finite-data regime, thus posing a challenge to learning good supports.
\citet{Kirichenko2020WhyNF} show that one can improve a \gls{mle}-trained normalizing flow architecture that assigns high likelihoods to \gls{ood} images by
directly minimizing the density of \gls{ood} images.
After showing the same for \acrshort{pixelcnn++} in \Cref{sec:pcnn_neg}, we show that in-distribution likelihoods can remain high even with the additional constraint of forcing down probabilities. This result suggests that existing model classes 
contain similarly good solutions which do not suffer from the problem of high likelihoods for particular \gls{ood} inputs; however, the combination of the model architecture coupled with existing gradient descent-based maximum likelihood optimization does not seem to find such solutions.

\section{Related Work}
\label{sec:related}

\paragraph{Likelihood Ratio Test Statistics.}
Given the poor results of \glspl{dgm} seen in \citet{Nalisnick2019DoDG}, several works propose likelihood ratio test statistics
\cite{Ren2019LikelihoodRF, Serr2020InputCA, Schirrmeister2020UnderstandingAD}. \citet{Ren2019LikelihoodRF} propose a likelihood ratio where the alternative distribution is the same \gls{dgm} model class trained on perturbed samples; \citet{Serr2020InputCA} use a distribution induced by a general image compressor; and \citet{Schirrmeister2020UnderstandingAD} train a \gls{dgm} on a general image distribution such as 80 Million Tiny Images.
Per the discussion in \Cref{sec:definition}, we can conclude that the optimal alternative depends on the set of out-distributions of interest.
\paragraph{Test Statistics Motivated by Typicality.}
\citet{DeepMind2019DetectingOI} devise a goodness-of-fit test based on how close the empirical entropy of a sample deviates from the empirical entropy of the training set. \citet{Morningstar2020DensityOS} learn a kernel density estimator or one-class SVM over multiple statistics jointly, similarly accounting for whether a test example's statistics deviate from values seen in the training set. \citet{Choi2018WAICBW} suggest a typicality test in the latent space of a normalizing flow but see better performance using a likelihood-based
test statistic which takes into account variance across an ensemble of \glspl{dgm}. 
\citet{Wang2020FurtherAO} learn a function to map training samples to a white noise sequence and classify as \gls{ood} any input that is not white noise after being transformed via this function. 
The empirical success of some of these test statistics has been described as evidence in favor
the need to take typicality into account,
but we offer an alternative conclusion in \Cref{sec:important}: these test statistics may compensate for a particular model misestimation.
\paragraph{Modifying the DGM.}
\citet{Kirichenko2020WhyNF} and \citet{Schirrmeister2020UnderstandingAD} both modify the invertible layers of a normalizing flow, improving \gls{ood} performance at the expense of worse likelihoods. \citet{BIVA} replace the posterior distributions of lower-level latents in their variational autoencoder with their priors, showing improved \gls{ood} performance but worse approximate likelihoods. 

\paragraph{Investigating Density for OOD Detection.}
\citet{Lan2020PerfectDM} 
suggest  that  density  may  be  limited  in  its  use for  anomaly  detection  
because
a transformation applied to a continuous random variable with strictly positive density can arbitrarily re-rank the density.
However, points outside of the support will still be outside of the support under transformations, meaning that such out-distribution 
samples cannot be
re-ranked to be higher than points in the in-distribution.

\paragraph{Test Statistics Based on Discriminative Models.} 
Methods based on discriminative models use some property of a learned classifier for discriminating in-distribution classes, such as the maximum softmax probability \cite{Hendrycks2017ABF}. Extensions incorporate \gls{ood} loss terms, such as encouraging the softmax probabilities of \gls{ood} samples to be uniform or utilizing temperature scaling to increase the sharpness of the in-sample probabilities \cite{Hendrycks2019DeepAD, Liang2018EnhancingTR}. These methods can be seen as learning a direct mapping from inputs to some \gls{ood} score based on minimizing risk with respect to a given distribution of in- and out-samples. However, 
these test statistics suffer from the same issue that motivates \gls{ood} detection in the first place: the learned conditional $\hat{p}(y \g \mbx)$ must perform extrapolation when the input is unlike what was seen during training, and even the true conditional $p(y \g \mbx)$ is not defined for inputs where $p(\mbx) = 0$.

\paragraph{Directly Learning a Decision Boundary.} One-class Support Vector Machines \cite{Schlkopf1999SupportVM}, Support Vector Data Description \cite{Tax2004SupportVD}, and
their deep variants 
\cite{Ruff2018DeepOC} learn to 
separate a subset of the input space with its complement. These methods estimate the distribution's support rather than the density or probability over that set.
Note that the support boundary is all that matters for detection if the family of relevant out-distributions only include those disjoint in support with the in-distribution. 

\section{Discussion}
The failures of existing \glspl{dgm} to detect certain out-distributions based on log-likelihood has 
prompted
some to wonder whether \gls{ood} detection based on probability models requires additional considerations in high dimensions. The results of our analysis suggest that it is the model that is at fault, not the method for \gls{ood} detection. We additionally highlight the importance of formalizing the out-distributions of interest for \gls{ood} detection in general, as well as the arbitrary choice of the typical set for \gls{ood} detection.

Understanding the \gls{ood} detection failures of \glspl{dgm} as estimation error introduces avenues for future work.
We 
suggest
that existing models are incorrectly assigning higher probability or density to certain natural images even when such images should have zero probability or density, and we hypothesize that the issue arises due to not just a single modeling choice, but the combination of the model architecture and maximum likelihood objective.

The extent of misestimation in existing \glspl{dgm} could be a relatively small amount of total probability mass, if the total volume of the out-distribution support (e.g. those of \acrshort{svhn}, \acrshort{mnist}) is relatively small in comparison to that of the in-distribution support (e.g. those of \acrshort{cifar-10}, Fashion\acrshort{mnist}). Under this scenario, a model could be near-perfect
yet still assign higher probability or density to samples from an out-distribution with disjoint support. This possibility illustrates the additional considerations required for \gls{ood} detection beyond for instance what is necessary for good generation or held-out likelihood.
That said, the bias exhibited by existing models may affect more than a negligible probability set (including if the problem exists for multiple out-distributions), meaning that future work directed towards correcting this bias across existing model classes could benefit not only \gls{ood} detection, but other applications for generative models as well. Further work comparing generative modeling and support detection may also provide insight. Finally, the interplay between \gls{ood} detection and epistemic uncertainty is worth further study, based on their shared relevance to predictive modeling.

\paragraph{Acknowledgements.}
This work was supported by NIH/NHLBI Award R01HL148248, NSF Award 1922658 NRT-HDR: FUTURE Foundations, Translation, and Responsibility for Data Science, NSF Award 1514422 TWC, and a DeepMind Fellowship. We thank the reviewers for their very helpful feedback and \citet{Kingma2018GlowGF} and \citet{Ren2019LikelihoodRF} for open-sourcing their code.

\bibliography{main.bbl}
\bibliographystyle{icml2021}

\appendix 
\onecolumn

\newcommand{\phiofx}{\phi_p(\mbx)}
\newcommand{\indinphi}{\indicator{\phiofx\in \Phi}}
\newcommand{\indnotinphi}{\indicator{\phiofx \notin \Phi}}
\newcommand{\pxgivenphi}{p(\mbx|\phiofx)} 
\newcommand{\qxgivenphi}{q(\mbx|\phiofx)}
\newcommand{\Aphi}{A_{\phiofx}}
\newcommand{\xinA}{\indicator{\mbx \in \Aphi}}
\newcommand{\xnotinA}{\indicator{\mbx \notin \Aphi}}

\section{Proposition 1}
\label{sec: prop1_proof}

\textit{Proposition} Let $P$ be the distribution under the null hypothesis $H_0$. Let $\mu$ be the measure associated with the distribution of test statistic $\phiofx$ under the null. Then, assuming conditional $\mbx \g \phiofx$ is not degenerate on $\mu$-non-measure zero set, there exists a set of alternative distributions $Q \in \mathcal{Q}$ where $Q \neq_d P$ and the test has power equal to the false positive rate. In other words, the test does no better than random guessing.

\begin{proof}
We first construct a distribution $q(\mbx) \neq_d p(\mbx)$ but where $q(\phiofx) = p(\phiofx)$.

The roadmap for this part of the proof is as follows:
for some function $f$, we write
\begin{align} 
\E_{p(\mbx)}(f_p) - \E_{q(\mbx)}(f_p) &= \E_{p(\phiofx)}\Big[\E_{\pxgivenphi}(f_p) - \E_{\qxgivenphi}(f_p)\Big]
\end{align} 
We then identify  $\qxgivenphi$ and $f_p$ such that the inner difference of expectations is non-zero, which implies inequality in distribution via $\E_{q(\mbx)}(f_p) \neq \E_{p(\mbx)}(f_p)$. We do not change the distribution in the outer expectation $p(\phiofx)$. We finally define 
$q(\mbx)=p(\phiofx)\qxgivenphi$.

We now show how to construct $f_p,q$. Let $(\Omega_{\phiofx}, \mathcal{F}_{\phiofx})$ be the probability space 
associated with $\phiofx$, with probability measure $\mu=\mathbb{P}_{p(\phiofx)}$.  By assumption, $p(\mbx \g \phiofx)$ is non-degenerate on some $\mu$ non-measure zero set. 
This means there exists a set $\Phi \in \mathcal{F}_{\phiofx}$ with $\mu(\Phi) > 0$ such that $ \forall \phiofx \in \Phi$, $\exists \Aphi \subset \text{supp}(p(\mbx \g \phiofx))$
such that $0<\mathbb{P}_{p(\mbx \g \phiofx)}(\Aphi)<1$.

Let $g$ be any function for which $\E_{\pxgivenphi}(g) < \infty$
$\forall \phiofx \notin \Phi$. Then define
\begin{align}
    f_{p}(\mbx) \triangleq \indinphi \xinA + \indnotinphi g(\mbx)
\end{align}
Define the conditional $\qxgivenphi$ with normalization constant $C_{\phiofx}$ and $0<\lambda < 1$:
\begin{equation} \label{eq:conditional}
\begin{split}
\qxgivenphi &\triangleq \indinphi \Big[\frac{1}{C_{\phiofx}}\Big(\lambda \pxgivenphi \xinA + \pxgivenphi \xnotinA\Big)\Big]\\
&\quad + \indnotinphi \pxgivenphi
\end{split}
\end{equation}
For $\phiofx \notin \Phi$, $\qxgivenphi = \pxgivenphi$.
Therefore, 
$\indnotinphi [\E_{\pxgivenphi}(f_p) - \E_{\qxgivenphi}(f)] = 0$. For 
$\phiofx \in\Phi$, $\qxgivenphi$ moves density away from points in $\Aphi$ relative to $\pxgivenphi$, given that $0 < \lambda < 1$. 

For simplicity, we construct $q$ such that $\text{supp}(q(\mbx \g \phiofx)) = \text{supp}(p(\mbx \g \phiofx))$. This is to avoid any issues with an invalid joint distribution $q(\mbx, \phiofx) \neq q(\mbx)$ if $q(\mbx | \phiofx) = 0$ (the left-hand side would be 0 while the right-hand side would be greater than 0 $\forall \mbx \in \text{supp}(p(\mbx))$.

We now show that this construction leads to $\E_{p(\mbx)}(f_p) - \E_{q(\mbx)}(f_p) > 0$, implying inequality in distribution.

\begin{align}
    &\E_{p(\mbx)}(f_p) - \E_{q(\mbx)}(f_p) \\
    = &\E_{ p(\phiofx)}\Big[\E_{\pxgivenphi}(f_{p}) - \E_{\qxgivenphi}(f_{p})\Big]\\
    = &\E_{p(\phiofx)}\indinphi \Big[\E_{\pxgivenphi}(f_{p}) - \E_{\qxgivenphi}(f_{p})\Big] \label{eq:indic} \\
    =&\E_{p(\phiofx)}\indinphi \Big[\E_{\pxgivenphi}(\xinA) - \E_{\qxgivenphi}(\xinA)\Big] \\
    =&\E_{p(\phiofx)}\indinphi \Big[\int_{\Aphi}\pxgivenphi d\mbx - \int_{\Aphi}\qxgivenphi d\mbx)\Big] \\ 
    =&\E_{p(\phiofx)}\indinphi \Big[\int_{\Aphi}\pxgivenphi d\mbx - \int_{\Aphi}\frac{1}{C_{\phiofx}}\lambda \pxgivenphi d\mbx)\Big] \label{eq:subs} \\
    > &0 \label{eq:end}
\end{align}

Line~\ref{eq:subs} follows from the substitution of $\qxgivenphi$ defined in \Cref{eq:conditional}. Line~\ref{eq:end} follows from the fact that $\frac{\lambda}{C_{\phiofx}} < 1$, shown below:
\begin{align}
    C_{\phiofx} &= \int_{\mathcal{X}} \lambda \pxgivenphi \xinA + \pxgivenphi \xnotinA d\mbx\\
    &= \lambda \mathbb{P}_{\pxgivenphi}(\Aphi) + \mathbb{P}_{\pxgivenphi}(\Aphi^C) \\
    \frac{\lambda}{C_{\phiofx}} &= \frac{\lambda}{\lambda \mathbb{P}_{\pxgivenphi}(\Aphi) + \mathbb{P}_{\pxgivenphi}(\Aphi^C)} \\
    &= \frac{1}{\mathbb{P}_{\pxgivenphi}(\Aphi) + \frac{1}{\lambda}\big[\mathbb{P}_{\pxgivenphi}(\Aphi^c)\big]} \\
    &< 1 \label{eq:end1}
\end{align}
Line~\ref{eq:end1} holds since the denominator in the previous line is greater than 1:
Since $0<\lambda < 1$, $\frac{1}{\lambda} > 1$. Then, $\mathbb{P}_{\pxgivenphi}(\Aphi) + \frac{1}{\lambda}
\Big[\mathbb{P}_{\pxgivenphi}(\Aphi^c)\Big] > \mathbb{P}_{\pxgivenphi}(\Aphi) + \mathbb{P}_{\pxgivenphi}(\Aphi^c) = 1$.

Having constructed the distribution $Q$, we now proceed with the second part of the proposition: for any specified false positive rate,
any test based on $\phi_p$ has power equal to the false positive rate when the \gls{ood} samples come from $Q$.

Recall that $q(\phiofx) = p(\phiofx)$. Then, for any rejection rule $\phiofx \not\in \Phi_{\text{Accept}}$, the probability of rejection is the same
regardless of whether the sample $\mbx$ is drawn from $P$ or $q$:
\begin{align}
    \forall \Phi_{\text{Accept}},
    \quad 
    \mathbb{P}_{\mbx \sim q}(\phiofx \not\in \Phi_{\text{Accept}}) &= 
    \mathbb{P}_{\mbx \sim p}(\phiofx \not\in \Phi_{\text{Accept}}).
\end{align}
Therefore, the power of the test (i.e. rejecting under the $H_A: \mbx \sim q$) is equal to the false positive rate (i.e. rejecting under $H_0: \mbx \sim p$).
When power and false positive rate are equal for all possible values of the false positive rate, then the result is an ROC curve $y = x$ with \gls{auc} $0.5$. This is equivalent to random guessing with rejection rate based on the false positive rate chosen for the test.
\end{proof}

\section{Rejection Rules Can Be Written in the Form $\phiofx < k$}  %
\label{sec:rej}

\paragraph{Lemma 1} \label{WLOG}\textit{Any rejection rule involving intervals, i.e. $\phi(\textbf{x}) \not\in \Phi$ can be recast as a rule of the form $\phi'(\textbf{x}) < k$.}

\textit{Proof} If we have a one-sided rule, i.e. an interval $\Phi$ where one of the endpoints is $-\infty$ or $\infty$, we simply reverse the sign if necessary, and for two-sided rules, i.e. a bounded interval, we can find the midpoint of the interval $m$, where $\Phi = [m - k, m + k]$, and recast the rule to $\lvert \phi(\textbf{x}) - m \rvert < k$. 

Rejection rules of this form match the same ``rejection'' rules used for binary classification more broadly. For added clarity, we define some \gls{ood} detection methods based on their rejection rules in this form. For instance, the likelihood-based test \cite{Bishop} rejects when the negative log likelihood is above a certain threshold $k$, whereas the typicality test \cite{DeepMind2019DetectingOI} rejects when the distance to the training set entropy is above $k$. 

\section{Details for Experiment 5.2}
\label{sec:decexp_supp}
In this experiment, we compare a partially trained \acrshort{glow} model $p_\mbtheta$ with a pretrained \acrshort{glow} model \cite{Kingma2018GlowGF} which we use as our data distribution $P$. First, we generate samples from $P$ by sampling from the  \acrshort{glow} model pretrained on \acrshort{cifar-10} \footnote{The pretrained model is available here: \href{https://openaipublic.azureedge.net/glow-demo/logs/abl-1x1-aff.tar}{https://openaipublic.azureedge.net/glow-demo/logs/abl-1x1-aff.tar}}. We use temperature 1 for sampling to ensure our samples come from the distribution specified by the model. We generate 40,000 samples for training and 10,000 samples for evaluation, matching the train and test set sizes of the \acrshort{cifar-10} dataset.

The \gls{glow} model $p_\mbtheta$ is  made of 3 blocks, each with 8 affine coupling layers with 400 hidden units per layer. The network is trained with Adamax at learning rate 0.001, which stays constant after 10 epochs of warmup. We use batch size 64 during training. We intentionally limit the training (50 epochs with 10 epochs of warmup) to make the model mis-estimation clear. Our model achieves an average bits per dimension of 3.67 on the test samples, versus 3.45 for the true model (lower is better). 

The true model is a larger model than $p_\mbtheta$, consisting of 3-blocks each with 32 affine coupling layers with 400 units each.

We evaluate \gls{ood} performance on the test set of the model samples and the test set of CelebA.

\section{Existing Model Architectures Can Yield Good OOD Detectors}
\label{sec:pcnn_neg}
We directly optimize a \acrshort{pixelcnn++} to distinguish between Fashion\acrshort{mnist} and \acrshort{mnist} by replacing the maximum likelihood training objective with one which simultaneously maximizes likelihood on Fashion\acrshort{mnist} images while minimizing likelihood of \acrshort{mnist} images. Our objective is similar to that of \citet{Kirichenko2020WhyNF}, who show that flows can distinguish problematic \gls{ood} dataset pairs when optimized directly to do so. 
\begin{equation} \label{eq:neg_obj}
    \frac{1}{N_{in}}\sum_{x \in \mathcal{D}_{in}} \log p_{\theta}(x) - \frac{1}{N_{ood}}\sum_{x' \in \mathcal{D}_{ood}} \min(\log p_{\theta}(x'), c)
\end{equation}

Replicating the architecture of \citet{Ren2019LikelihoodRF}, we train our model across five random seeds using the MLE objective and five seeds with the above objective. We use the same training hyperparameters as \citet{Ren2019LikelihoodRF}: 50,000 steps at a learning rate of 0.0001 with exponential decay rate of 0.999995 per step, batch size of 32, and Adam optimizer with momentum parameters 0.95 and 0.9995. Our results, shown in \Cref{table:pcnn_neg}, demonstrate that the \acrshort{pixelcnn++} architecture has the capacity to push down probabilities on problematic \gls{ood} samples while maintaining high in-distribution likelihoods.

\begin{table}[ht] 
\centering
\caption{ \label{tab:neg_train} Existing models can be optimized to distinguish datasets. \acrshort{pixelcnn++} trained via the negative training (NT) objective in \Cref{eq:neg_obj} can achieve near-perfect \gls{ood} detection while maintaining comparable held-out log likelihoods (LL) with models trained via maximum likelihood estimation (MLE). We report mean and standard deviation of the results over 5 random seeds.}
\begin{tabular}{rrrr}
\toprule
 & Fashion LL & \gls{ood} \acrshort{auc} \\
 \midrule
 MLE &$ -1550 \pm$ 6 & $0.097 \pm 0.004$ \\
 NT & $-1562 \pm$ 7 & $1.000 \pm 0.000$ \\
\bottomrule
\label{table:pcnn_neg}
\end{tabular}
\end{table}

\end{document}